# Multiple Manifolds Metric Learning with Application to Image Set Classification

Rui Wang[a], Xiao-Jun Wu[a*], Kai-Xuan Chen[a], Josef Kittler[b]
[a]School of Internet of Things Engineering, Jiangnan University, Wuxi 214122, China
[b]CVSSP, University of Surrey, Guildford, GU2 7XH, UK
{cs_wr, wu_xiaojun}@jiangnan.edu.cn, kaixuan_chen_jnu@163.com, j.kittler@surrey.ac.uk

*Abstract*—In image set classification, a considerable advance has been made by modeling the original image sets by second order statistics or linear subspace, which typically lie on the Riemannian manifold. Specifically, they are Symmetric Positive Definite (SPD) manifold and Grassmann manifold respectively, and some algorithms have been developed on them for classification tasks. Motivated by the inability of existing methods to extract discriminatory features for data on Riemannian manifolds, we propose a novel algorithm which combines multiple manifolds as the features of the original image sets. In order to fuse these manifolds, the well-studied Riemannian kernels have been utilized to map the original Riemannian spaces into high dimensional Hilbert spaces. A metric Learning method has been devised to embed these kernel spaces into a lower dimensional common subspace for classification. The state-of-the-art results achieved on three datasets corresponding to two different classification tasks, namely face recognition and object categorization, demonstrate the effectiveness of the proposed method.

## I. INTRODUCTION

Recently, image set classification has proved to be an area of increasing vitality in the domain of computer vision [1, 2, 3, 4, 5, 6] due to its flexibility and effectiveness of feature representation as compared with single image classification. In image set classification, each set generally contains a number of images that belong to the same class, and most of which exhibit a wide range of rigid and non-rigid variations as well as illumination changes. The selection of both gallery and probe is always arbitrary. In other words, we can choose one or more sets of each category for training and the rest for testing. In the era of big data, the increased attractiveness of image set classification naturally manifests itself in biometric applications including video-based face recognition, surveillance, person re-identification, and bioinformatics certification. The success of these applications proves that the image set can provide more discriminatory information than single image.

When conducting image set classification, one of the most important steps is to model the original image set. The second-order statistics, i.e. covariance matrix and linear subspace are two widely used models which dues to their flexibility, simplicity, and sufficiency to characterize the distribution of the given image set. According to previous studies [7, 8, 9], the specific geometry spanned by covariance matrix, or linear subspace, is the non-linear Riemannian manifold. Specifically, they are the SPD manifold and Grassmann manifold respectively. Another important step is to measure the similarity between any two image sets. However, the conventional feature learning algorithms based on Euclidean space cannot be applied to the Riemannian manifold directly. To overcome this obstacle, [10, 11, 12, 13] introduced some distance metrics, such as Log-Euclidean Distance (LED) [11], Projection Metric (PM) [13], Affine-Invariant Riemannian Metric (AIRM) [10], and Stein divergence [12] for Riemannian manifold with the purpose of encoding the non-Euclidean geometry properly.

By utilizing these well-defined Riemannian metrics, some Riemannian manifold learning methods try to map the original space into a high dimensional Hilbert space via a kernel function with respect to the Riemannian distance metric [11, 13, 10, 12], and then learn a map to transform it to a low dimensional one. This computing strategy enables the original manifold to be reduced to an approximate Euclidean space, where Euclidean based computation algorithms can be applied for classification. Nevertheless, the high computational complexity and no consideration of the manifold property are the main shortcomings of these methods. Recently, some algorithms that jointly perform metric learning and dimensionality reduction directly on Riemannian manifold have been suggested [9, 14, 15]. The generated lower dimensional Riemannian manifold, which is more discriminative and computationally efficient, as well as being cognizant of the manifold property are the main advantages of these algorithms. However, as the linear transformation process is executed on a non-linear manifold, it inevitably leads to sub-optimal results.

In order to improve the classification performance on complicated datasets, some authors extended the idea of conventional deep neural networks to Riemannian manifold, and applied the Riemannian manifold stochastic gradient descent technique to learn the parameters [16, 17]. Unfortunately, the improvement of recognition accuracy is at the expense of unacceptably high costs of training. Although existing image set classification methods have produced good results in some application areas, they pose many challenges as well. The main challenge is how to efficiently model the image-set without losing discriminative information.

In this paper, we propose an efficient method for image set classification. For each given image set, we first use the covariance matrix and linear subspace to model it. Based on the well-studied Riemannian metrics, we adopt the kernel method to embed the original Riemannian manifold into Reproducing Kernel Hilbert Space (RKHS). In order to enhance inter-class dispersion and intra-class compactness, a metric learning algorithm is developed to learn a distance metric for novel features extracted from the multiple manifolds. Extensive classification experiments on three different benchmark datasets, validate the proposed method, and demonstrate its effectiveness.

## II. RELATED WORK

In image set classification, traditional methods can be divided into four kinds: kernel based discriminative learning methods, subspace based learning methods, manifold dimensionality reduction methods, and deep learning methods. The key advantages of linear subspace modeling of image sets are lower computational complexity and high discriminatory ability of the learned features. Manifold-to-Manifold Distance (MMD) [18] is an efficient subspace learning algorithm which clusters each image set into multiple linear local models. This simplifies the computation of the manifold distance to a distance defined on the subspace. As a result, the distance computing is easy and precise. However, the process of establishing the local models is expensive. When the datasets have large between-class variations, the subspace based methods cannot characterize the complex data distribution information very well. Fortunately, some kernel based methods have been proposed to map the original Riemannian space into Hilbert space where discriminative learning is conducted for classification. Wang et al. in [7] derived a kernel function which is based on Log-Euclidean distance (LED) [11] to project the SPD matrices from the SPD manifold to an approximate Euclidean space, and then the Kernel Discriminant Analysis (KDA) [19] algorithm is applied for further feature learning. This method improves the classification scores on some complicated datasets, but it also ignores the manifold property of the data.

Recently, some new approaches [9, 14, 15] which aim to directly perform metric learning and dimensionality reduction on Riemannian manifold have been proposed. They try to transfer the high-dimensional Riemannian manifold into a lower-dimensional, more discriminative one without losing the Riemannian manifold geometry. Harandi et al. in [15] has put forward a method, which first makes use of the intra-class similarity and inter-class dissimilarity to build the affinity matrix between any two SPD matrices. Then the Affine-Invariant Riemannian Metric (AIRM), Stein divergence, and Log-Euclidean distance (LED) can be used to construct the discriminant function, respectively. Lastly, the Riemannian Conjugate Gradient (RCG) [20] optimization algorithm is adopted to find the target transformation matrix. The performance achieved on different tasks indicates the reliability of this method. However, the optimization process is time consuming in the training stage. Furthermore, the classification results may be sub-optimal owing to the fact that the linear mapping process is implemented on a non-linear manifold.

More recently, some authors extended the conventional deep learning architectures to Riemannian manifold, in order to perform non-linear feature learning, and also created some excellent works for SPD matrix learning and subspace learning [16, 17]. From the foregoing description, we can see that the single modeling options for any given image set are widely available. The Hybrid Euclidean-and-Riemannian Metric Learning (HERML) [21] is a method that combines multiple heterogeneous statistics, such as mean, covariance, and Gaussian distribution for image set classification. As different statistics features lie in different spaces, a metric learning algorithm is devised to fuse them into a common subspace to improve classification. This method is robust in complex classification tasks, but computationally complex.

## III. PROPOSED METHOD

An overview of the proposed algorithm is given in Fig. 1. We first transfer the given image set into SPD manifold and Grassmann manifold using the covariance matrix and linear subspace for set modeling, respectively. For each extracted feature distribution, we use the kernel method to remap it to a high dimensional Hilbert space. In order to improve the separability of different classes on the manifolds, a distance metric learning is performed in the kernel space. Finally, the nearest neighbor (NN) classifier is applied for classification.

### A. Set Modeling with Multiple models

Let $S_i = [s_1, s_2, ..., s_{n_i}]$ be an image set with $n_i$ images, where $s_i \in R^{d \times 1}$ represents the $i$-th image sample. Given each original image set, we regard the SPD manifold and Grassmann manifold, which are spanned by covariance matrices and linear subspaces, as the extracted new data distributions. They can therefore used as feature representations of image set.

**The Geometry of SPD manifold**: For all non-zero $v \in R^{d \times 1}$, a real Symmetric Positive Definite (SPD) matrix $C \in R^{d \times d}$ has the property of $v^T C v > 0$. We use $S_{++}^d$ to denote the space of the interior of a convex cone of the SPD matrix $C$ in the $d(d+1)/2$-dimensional Euclidean space. It is a specific SPD manifold as studied in [10, 11], and $C$ can be computed by:

$$C = \frac{1}{n-1} \sum_{i=1}^{n} (s_i - m)(s_i - m)^T \quad (1)$$

where $m$ is the mean vector of a given set. In order to form a valid SPD manifold, the positive definite property of $C$ is regularized by:

$$C^* = C + \frac{tr(C)}{\alpha} I_d \quad (2)$$

where $I_d$ is an identity matrix. We let $\alpha = 10^3$ in all the experiments.

When considering the distance metric on $S_{++}^d$, the widely used Log-Euclidean Distance (LED) [11] is adopted to measure the geodesic distance. It is expressed as:

$$d(C_i, C_j) = \|\log(C_i) - \log(C_j)\|_F \qquad (3)$$

where $C_i, C_j \in S_{++}^d$, log is the matrix logarithm operator, and $\|\cdot\|_F$ represents the matrix Frobenius norm. Under the metric, the SPD manifold simply reduces to a flatten space $T_I$, and a Riemannian kernel function can be derived by computing the inner product as [7, 22],

$$k_{\log}(C_i, C_j) = tr[\log(C_i)\log(C_j)] \qquad (4)$$

the validity of which has been proven in [7].

**The Geometry of Grassmann manifold**: A Grassmann manifold $G(q,d)$ is spanned by a set of linear subspaces of $R^{d\times q}$. Each linear subspace, spanned by an orthonormal basis matrix $Y \in R^{d\times q}$ with the constraint $Y^T Y = I_q$, can be regarded as an element of $G(q,d)$. After applying singular value decomposition (SVD) for the covariance matrix $C$, s.t. $C = U\Sigma U^T$, where $U$, $\Sigma$ correspond to the eigenvectors and eigenvalues respectively, each input image set can be represented by $q$ largest eigenvectors of $C$, which form the orthonormal basis matrix $Y$, specifically, a linear subspace.

As is noted in [23], each point that resides on the Grassmann manifold corresponds to a unique projection matrix $YY^T$, which can be used to re-represent the Grassmann manifold element. As proved in [23], the projection metric can give a precise approximation for the true geodesic distance on $G(q,d)$. For any pair of projection operators $Y_1Y_1^T$, $Y_2Y_2^T$, this metric is formulated as:

$$d(Y_1Y_1^T, Y_2Y_2^T) = 2^{-1/2} \|Y_1Y_1^T - Y_2Y_2^T\|_F \qquad (5)$$

A projection kernel can be induced by:

$$k_p(Y_1Y_1^T, Y_2Y_2^T) = tr[(Y_1Y_1^T)(Y_2Y_2^T)] = \|Y_1^T Y_2\|_F^2 \qquad (6)$$

It has been proved to be a well-defined Grassmann kernel [6].

*B. Multiple Manifolds Metric Learning*

Let $T = [S_1, S_2, \ldots S_N]$ be the gallery composed by $N$ image sets, with each set expressed as: $S_i = [s_1, s_2, \ldots, s_{n_i}] \in R^{d\times n_i}$, where $1 \le i \le N$, and $n_i$ is the number of images in the $i$-th image set, and $S_i$ belongs to category $l_i$. For the kernel method, its non-linear mapping process can be described as: $\phi: M \to F$, $S \to \phi(S)$, where M is the original data space and $\phi(S)$ represents the new feature representation of $S$ in new space $F$. Though the mapping function $\phi$ is often implicit, kernel trick method [19] can be used to solve this problem. For simplicity, we first assume $\phi$ as an explicit mapping function. Hence, we denote $\phi_i^q$ as the extracted high dimensional feature of $q$-th model with respect to a given image set $S_i$. Here, $1 \le q \le Q$, and in this paper $Q = 2$ as we use two different models. Now, given any two training image sets $S_i$ and $S_j$, the distance metric defined in space $F$ can be written as:

$$d(S_i, S_j) = tr(\sum_{q=1}^{Q} u_q (\phi_i^q - \phi_j^q)^T P (\phi_i^q - \phi_j^q) u_q) \qquad (7)$$

where $u_q$ is the added connection weight for each selected model, which is initialized by handcrafting and determined by cross validation. $P$ is the Mahalanobis matrix which needs to be learned, and since $P$ is symmetric positive semi-definite (SPSD), we can look for a non-square matrix $W = [w_1, w_2, \ldots w_{d_z}]$ to reconstruct $P$: $P = WW^T$. So that, the Eq.7 can be rewritten as:

$$d(S_i, S_j) = tr[W^T (\sum_{q=1}^{Q} u_q (\phi_i^q - \phi_j^q)(\phi_i^q - \phi_j^q)^T u_q) W] \qquad (8)$$

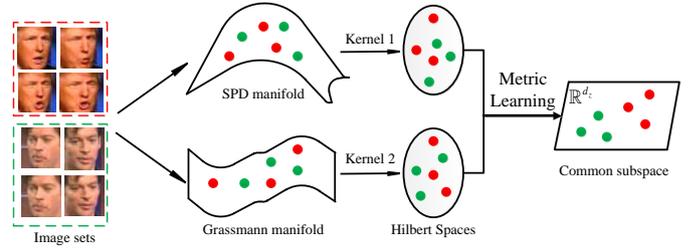

Fig. 1. Conceptual illustration of the proposed image set classification framework

The discriminant function can therefore be designed for the purpose of maximizing the within-class compactness and the between-class distance. The target matrix $W$ is thus obtained by solving the following objective function :

$$W^* = \arg\max_W J(W) = \arg\max_W \frac{J_b(W)}{J_w(W)} \qquad (9)$$

where $J_w(W)$, $J_b(W)$ indicate the intra-class compactness and inter-class dispersion respectively, and can be formulated as:

$$J_w(W) = \frac{1}{M_w} \sum_{i=1}^{N} \sum_{j: l_i = l_j} d(S_i, S_j) = tr(W^T R_w W) \qquad (10)$$

$$J_b(W) = \frac{1}{M_b} \sum_{i=1}^{N} \sum_{j: l_i \ne l_j} d(S_i, S_j) = tr(W^T R_b W) \qquad (11)$$

where $M_w$ is the number of sample pairs from the same class, $M_b$ is number of sample pairs from different classes. $R_w$ represents the within-class scatter, $R_b$ represents the between-class scatter. The specific form of them are as follow:

$$R_w = \frac{1}{M_w} \sum_{i=1}^{N} \sum_{j:l_i=l_j} \sum_{q=1}^{Q} u_q (\phi_i^q - \phi_j^q)(\phi_i^q - \phi_j^q)^T u_q \quad (12)$$

$$R_b = \frac{1}{M_b} \sum_{i=1}^{N} \sum_{j:l_i \neq l_j} \sum_{q=1}^{Q} u_q (\phi_i^q - \phi_j^q)(\phi_i^q - \phi_j^q)^T u_q \quad (13)$$

Indeed, it is very difficult for us to compute $R_w$ and $R_b$ for the reason of the implicit mapping function $\phi$. However, we regard the basis matrix $w_h$ as a linear combination of all the training samples in space $F$, i.e.,

$$w_h = \sum_{i=1}^{N} e_i^h \phi_i^q \quad (14)$$

where $e_i^h$ are the representation coefficients. So we can have,

$$\sum_{q=1}^{Q} w_h^T \phi_i^q = \sum_{i=1}^{N} \sum_{q=1}^{Q} e_i^h (\phi_i^q)^T \phi_i^q = \sum_{q=1}^{Q} (e^h)^T K_{\cdot i}^q \quad (15)$$

where $e^h \in R^{N \times 1}$ is a column vector and $e_i^h$ denotes its $i$-th entry. $K_{\cdot i}^q$ is the $i$-th column of the $q$-th kernel matrix $K^q \in R^{N \times N}$, which is generated from the $q$-th model by using the corresponded $q$-th Riemannian kernel.

Hence, Eq.9, Eq.12, and Eq.13 can be rewritten as:

$$E^* = \arg\max_E J(E) = \arg\max_E \frac{tr(E^T R_b^* E)}{tr(E^T R_w^* E)} \quad (16)$$

where

$$R_w^* = \frac{1}{M_w} \sum_{i=1}^{N} \sum_{j:l_i=l_j} \sum_{q=1}^{Q} u_q (K_{\cdot i}^q - K_{\cdot j}^q)(K_{\cdot i}^q - K_{\cdot j}^q)^T u_q \quad (17)$$

$$R_b^* = \frac{1}{M_b} \sum_{i=1}^{N} \sum_{j:l_i \neq l_j} \sum_{q=1}^{Q} u_q (K_{\cdot i}^q - K_{\cdot j}^q)(K_{\cdot i}^q - K_{\cdot j}^q)^T u_q \quad (18)$$

**Optimization.** The optimization (trace of ratio) problem (16) can be transformed into the corresponding ratio of trace problem [24], which is simpler:

$$E^* = \arg\max_E Tr[(E^T R_w^* E)^{-1} (E^T R_b^* E)]$$
$$= \arg\max_E \frac{|E^T R_b^* E|}{|E^T R_w^* E|} \quad (19)$$

which can be directly solved by applying eigen-decomposition to $(R_w^*)^{-1} R_b^*$, and choosing its $d_z$ largest eigenvectors to compose the target transformation matrix $E = [e_1, e_2, ..., e_{d_z}]$.

## IV. EXPERIMENTS

To evaluate the performance of the proposed algorithm, we experiment on two different tasks: video-based face recognition and set-based object categorization. For the face recognition, we utilize the widely used YouTube Celebrities (YTC) [7, 8, 9, 14] dataset and Honda/UCSD dataset [7, 18]. The benchmark dataset ETH-80 [7, 6, 9, 14] is employed for the object categorization.

### A. Dataset introduction and experimental settings

The challenging dataset YouTube Celebrities (YTC) consists of 1910 video clips of 47 subjects that are collected from YouTube. Each clip consists of hundreds of frames, most of which are low resolution and highly compressed with noise and low quality. The number of image sets in each subject is not fixed. In our experiment, the size of each image is resized to $20 \times 20$ and we extract its grayscale feature for subsequent computation.

The dataset of ETH-80 is composed by 8 categories with each category containing 10 objects. There are 41 images of different views in each image set. The size of each image is $256 \times 256$. In order to be consistent with the existing literature, we also resize them to $20 \times 20$ and exploit the grayscale information of each image.

The Honda/UCSD dataset consists of 59 video sequences recording 20 different people. Each video sequence is composed of more than one hundred frames, most of which exhibit large variations in head pose and facial expression. For a fair comparison, we extract the grayscale feature of each image which has been resized to $20 \times 20$.

We conducted ten-fold cross validation experiments and ten randomly selected gallery/probe combinations, which follows the same protocol as some existing works [6, 7, 9, 14, 15, 21]. The classification results are reflected by the average recognition rates. As to YTC, each person had three randomly selected image sets for training and six for testing. For ETH-80, we randomly chose five image sets in each category for gallery and the other five for probes. However, in Honda dataset, we randomly chose one image set in each class to compose the gallery, and the rest was used for probes.

### B. Comparative methods

In order to study the effectiveness of the proposed method, we compare our method with existing algorithms including Discriminant Canonical Correlation (DCC) [25], Manifold-to-Manifold Distance (MMD) [18], Covariance Discriminant Learning (CDL) [7], Grassmann Discriminant Analysis (GDA) [6], Grassmannian Graph-Embedding Discriminant Analysis (GGDA) [26], Projection Metric Learning (PML) [14], Log-Euclidean Metric Learning (LEML) [9], SPD Manifold Learning based on stein divergence (SPDML-Stein) [15] and affine-invariant metric (SPDML-AIM) [15], Localized Multi-Kernel Metric Learning (LMKML) [27], and Hybrid

TABLE I. AVERAGE CLASSIFICATION ACCURACIES (%) OF DIFFERENT METHODS ON HONDA DATASET

| Method | DCC [25] | CDL [7] | PML [14] | LEML [9] | LMKML [27] | **MMML** |
|---|---|---|---|---|---|---|
| Honda | 94.75±1.32 | 98.13±2.64 | 98.44±2.21 | 98.75±2.19 | 98.50±2.14 | **99.06±2.11** |

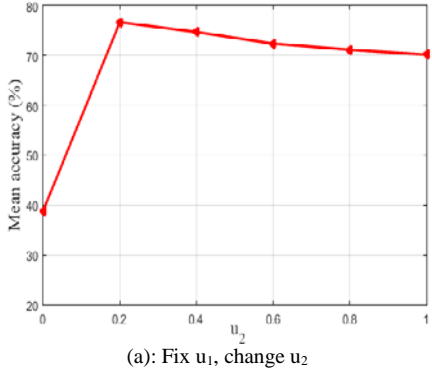

(a): Fix $u_1$, change $u_2$

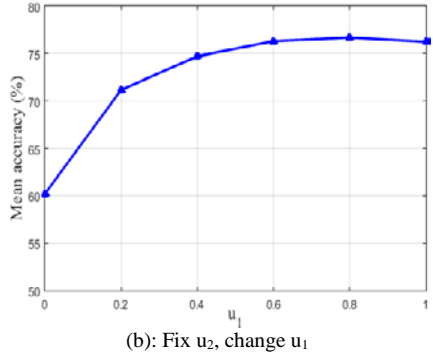

(b): Fix $u_2$, change $u_1$

Fig. 2. The impact of the connection coefficient $u_q$ on the final classification accuracy on YTC dataset.

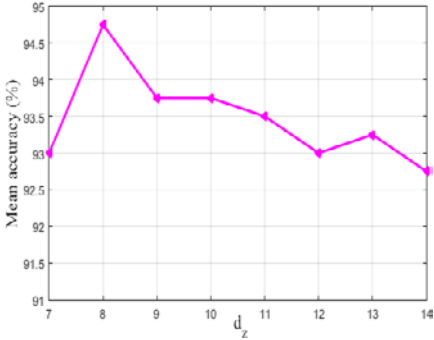

Fig. 3. The impact of the subspace dimension $d_z$ on the final classification accuracy on ETH-80 dataset.

Euclidean-and-Riemannian Metric Learning (HERML) [21]. Both GDA and GGDA are Grassmannian kernel learning methods. GGDA utilizes a discriminative graph embedding framework to improve the robustness and discriminatory ability of GDA. When it comes to Riemannian manifold dimensionality reduction problem, PML, LEML, and SPDML could be the most representative methods, which jointly perform metric learning and dimensionality reduction, for the purpose of better preserving the Riemannian geometry. To improve the classification performance on the complicated datasets, LMKML and HERML exploit the multiple statistics features to model the given image set, and devise a metric learning framework to better fuse the heterogeneous data spaces.

TABLE II. AVERAGE CLASSIFICATION ACCURACIES (%) OF DIFFERENT METHODS ON YTC AND ETH-80 DATASETS

| Methods | YTC | ETH-80 |
|---|---|---|
| MMD [18] | 62.90 ± 3.24 | 85.72 ± 8.29 |
| DCC [25] | 65.48 ± 3.51 | 90.75 ± 4.42 |
| GDA [6] | 65.78 ± 3.34 | 93.25 ± 4.80 |
| GGDA [26] | 66.37 ± 3.52 | 94.32 ± 2.40 |
| CDL [7] | 68.76 ± 2.96 | 93.75 ± 3.43 |
| PML [14] | 67.62 ± 3.32 | 90.00 ± 3.54 |
| LEML [9] | 69.04 ± 3.84 | 92.25 ± 2.19 |
| SPDML-AIM [15] | 64.66 ± 2.92 | 90.75 ± 3.34 |
| SPDML-Stein [15] | 61.57 ± 3.43 | 90.50 ± 3.87 |
| LMKML [27] | 70.31 ± 2.52 | 90.00 ± 4.08 |
| HERML [21] | 74.60 ± 3.34 | 94.50 ± 3.37 |
| **MMML** | **76.70 ± 2.81** | **95.00 ± 1.89** |

We should emphasize that the reported results for DCC, MMD, SPDML, GGDA, and HERML on the three datasets were obtained by the original authors. We have our own implementation of CDL, GDA, PML and other comparative methods. The essential parameters of these methods were empirically tuned according to the recommendations in the original references. The average recognition scores on YTC and ETH-80 datasets are listed in Tab. 2. Tab. 1 shows the average classification results on Honda/UCSD dataset.

*C. Results and discussion*

Tab. 1 and Tab. 2 report the recognition rates of different comparative methods on the different datasets. From the results, we can intuitively see our method shows better performance both in terms of recognition scores and standard derivations. It is also interesting to find our method achieves a significant improvement in recognition rate on the much challenging YTC dataset. As to CDL, GDA, and GGDA, a common principle shared by them is the kernel method. However, these algorithms ignore the manifold geometry, and thus produce inferior results. The classification results produced by PML, LEML, and SPDML are also sub-optimal, due to the linear mapping function being learned on a non-linear manifold. We find when we utilize the multi-order statistics to extract features from a given image set, the classification performance of LMKML and HERML is better than that of the others, because the different statistics provide complementary feature information. Nevertheless, as the different extracted features reside in heterogeneous spaces, their fusion is inappropriate.

Note, the coefficient $u_q$ plays a pivotal role in the proposed algorithm. In order to optimize its value, we conducted experiments on the YTC dataset to evaluate its impact on the classification results. We first fix $u_1$ to 1, to determine the best value of $u_2$, and then $u_2$ is fixed, to search the best value of $u_1$. The final observations are shown in Fig. 2, and the best values of $u_1$ and $u_2$ that we chose are 0.8 and 0.2 respectively. Due to space limitation, the detailed procedure of determining

$u_1$ and $u_2$ on the other two datasets is not elaborated. However, the best values are the same as with the YTC dataset.

The purpose of introducing metric learning method in the proposed framework is to better integrate heterogeneous feature information in the resulting common subspace. As different $d_z$ may bring about different classification results, we experimented on ETH-80 dataset to verify its effect. In Fig. 3 we show the subspace dimension is surprisingly low.

## V. CONCLUSION

In this paper, we propose a novel image set classification algorithm, which can better integrate different Riemannian manifolds together by using a metric learning method. We evaluate this proposed algorithm on three different benchmark datasets, which involve two classification tasks. The results of extensive experiments show that the proposed method outperforms the state-of-the-art methods in terms of classification results and robustness. In further work, we plan to integrate other Riemannian manifolds into this framework to make the extracted features more discriminative, and to develop other metric learning algorithms.

ACKNOWLEDGMENT

The paper is supported by the National Natural Science Foundation of China (Grant No.61373055，61672265), UK EPSRC Grant EP/N007743/1, MURI/EPSRC/dstl Grant EP/R018456/1, and the 111 Project of Ministry of Education of China (Grant No. B12018).